\documentclass[sn-mathphys-num]{sn-jnl}

\usepackage{graphicx}%
\usepackage{multirow}%
\usepackage{amsmath,amssymb,amsfonts}%
\usepackage{amsthm}%
\usepackage{mathrsfs}%
\usepackage[title]{appendix}%
\usepackage{xcolor}%
\usepackage{textcomp}%
\usepackage{manyfoot}%
\usepackage{booktabs}%
\usepackage{algorithm}%
\usepackage{algorithmicx}%
\usepackage{algpseudocode}%
\usepackage{listings}%

\usepackage{threeparttable}
\usepackage{latexsym}
\usepackage{color}
\usepackage{enumerate}
\usepackage{subfigure}
\usepackage{diagbox}
\usepackage{colortbl}
\usepackage{bm}
\usepackage{makecell}
\usepackage{enumitem}
\usepackage{footnote}

\definecolor{mygray}{gray}{0.75}
\definecolor{darkgreen}{RGB}{34,139,34}
\newcommand{\green}[1]{\textcolor{darkgreen}{#1}}
\newcommand{\bb}[1]{\textcolor{cyan}{#1}}
\newcommand{\gbf}[1]{\green{\bf{#1}}}

\theoremstyle{thmstyleone}%
\theoremstyle{thmstyletwo}%

\theoremstyle{thmstylethree}%

\begin{document}

\title{StarLKNet: Star Mixup with Large Kernel Networks
for Palm Vein Identification}


\author[1,]{\fnm{Xin} \sur{Jin}}\email{jinxin4@ctbu.edu.cn}
\equalcont{These authors contributed equally to this work.}

\author[1,]{\fnm{Hongyu} \sur{Zhu}}\email{zhuhongyu@ctbu.edu.cn}
\equalcont{These authors contributed equally to this work.}

\author[2]{\fnm{Mounîm} \sur{A. El Yacoubi}}

\author[1]{\fnm{Haiyang} \sur{Li}}

\author[1]{\fnm{Hongchao} \sur{Liao}}

\author[1]{\fnm{Huafeng} \sur{Qin}}

\author*[1,]{\fnm{Yun} \sur{Jiang}}\email{jiang1@ctbu.edu.cn}

\affil[1]{\orgname{Chongqing Technology and Business University}, \orgaddress{\city{Chongqing}, \country{China}}}

\affil[2]{\orgname{Telecom SudParis, Institut Polytechnique de Paris}, \orgaddress{\city{Paris}, \country{France}}}


\abstract{As a representative of a new generation of biometrics, vein identification technology offers a high level of security and convenience.
Convolutional neural networks (CNNs), a prominent class of deep learning architectures, have been extensively utilized for vein identification.
Since their performance and robustness are limited by small \emph{Effective Receptive Fields} (\emph{e.g.}, 3$\times$3 kernels) and insufficient training samples, however, they are unable to extract global feature representations from vein images effectively.
To address these issues, we propose \textbf{StarLKNet}, a large kernel convolution-based palm-vein identification network, with the Mixup approach.
Our StarMix learns effectively the distribution of vein features to expand samples. To enable CNNs to capture comprehensive feature representations from palm-vein images, we explored the effect of convolutional kernel size on the performance of palm-vein identification networks and designed LaKNet, a network leveraging large kernel convolution and gating mechanism. In light of the current state of knowledge, this represents an inaugural instance of the deployment of a CNN with large kernels in the domain of vein identification.
Extensive experiments were conducted to validate the performance of StarLKNet on two public palm-vein datasets. The results demonstrated that \textbf{StarMix} provided superior augmentation, and \textbf{LakNet} exhibited more stable performance gains compared to mainstream approaches, resulting in the highest identification accuracy and lowest identification error.}




\maketitle

\section{Introduction}\label{sec1}

The issue of personal information security has received increasing attention in modern society, as misidentification may have a catastrophic impact on personal property security and privacy. Token-based authentication methods such as passwords and ID cards are at risk of being forgotten or stolen. In recent decades, there has been a great deal of research conducted on biometrics technology, based on the identification of individuals through their physiological(\emph{e.g.}, face  \cite{Liu2018faceide}, fingerprint \cite{Chugh2018fingerprint} and vein \cite{Wu2023adversarial, Yang2022cgan}) or behavioral(\emph{e.g.}, gait \cite{Fan2023opengait} and eye movement \cite{Qin2024emmixformer}) characteristics. The most biometric features used in applications are faces and fingerprints. However, these external features may be subject to potential forgery attacks \cite{Mathur2020IntroducingRO}. In contrast, the advantages of vein identification are significant. Veins and blood vessels are located inside the human body and are not easily affected by the external environment(\emph{e.g.}, skin moisture and wear). Furthermore, vein identification technology has inherent liveness detection since deoxygenated hemoglobin only exists in the living body. 
Methods based on deep learning can achieve end-to-end feature extraction without prior assumptions, but the training of network parameters requires a large amount of data support. Unfortunately, it is challenging to obtain a substantial number of samples from each class in practical applications due to limited storage and privacy policies. How to train a robust high-performance network with limited data is a pressing problem.
\begin{figure*}[t]
    \centering
    \includegraphics[width=1.0\linewidth]{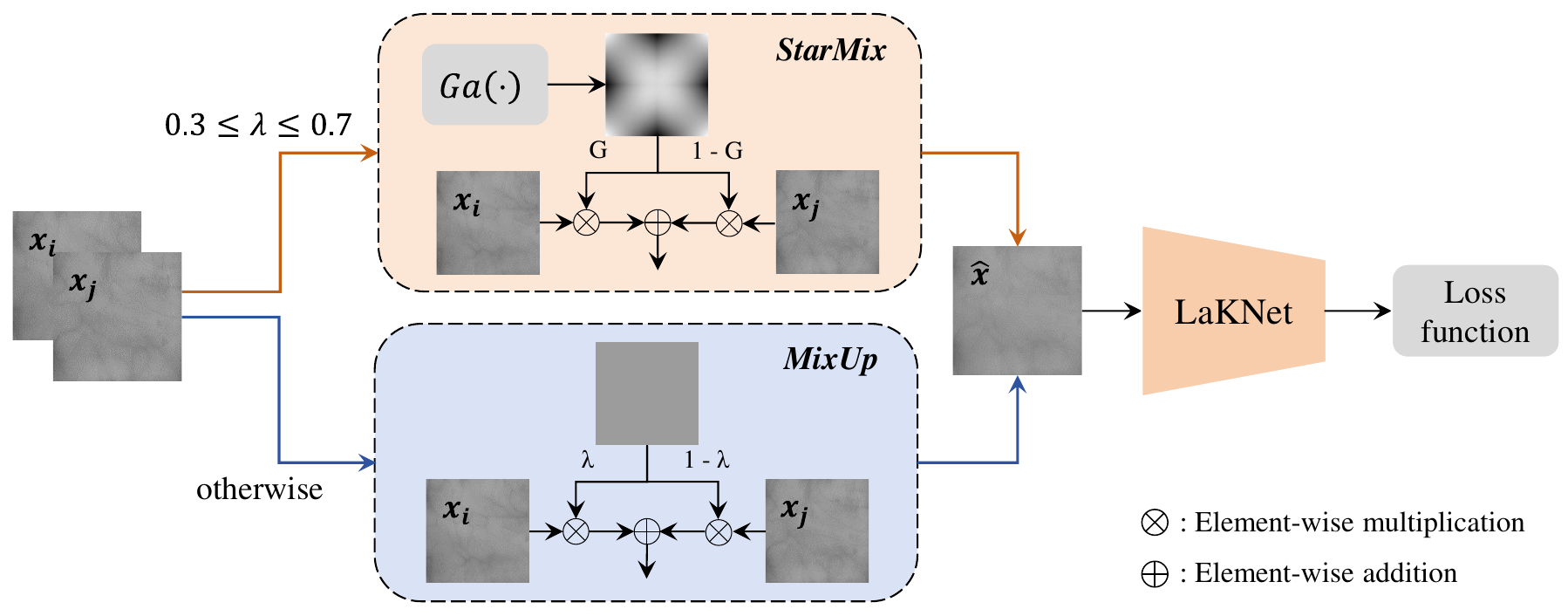}
    \caption{StarLKNet training workflow. After getting the mixing ratio, choose different mixing methods according to the threshold setting, get the mixed samples according to the different mixing methods, and then go through the encoder to get the final prediction and then calculate the loss to complete the training once.}
    \label{fig: StarMix}
\end{figure*}

To address the overfitting issue arising from insufficient training data, researchers have proposed data augmentation (DA) methods, through either hand-crafted or generative ways to generate new data for expanding the training set. Mixup uses linear interpolation to mix two or more samples. It has been widely utilized and improved by researchers, thanks to its plug-and-play functionality and minimal additional time overhead. Mixup has demonstrated outperformance and robust generalizability in downstream tasks, \emph{e.g.}, Super-resolution \cite{Yoo2020cutblur}, Segmentation \cite{Olsson2021classmix}, Regression \cite{Yao2022c}, and Long-tail \cite{Chou2020remix}. Another challenge for vein identification is
\begin{wrapfigure}{r}[0cm]{0pt} 
    \label{fig:acc}
    \centering
    \includegraphics[scale=0.24]{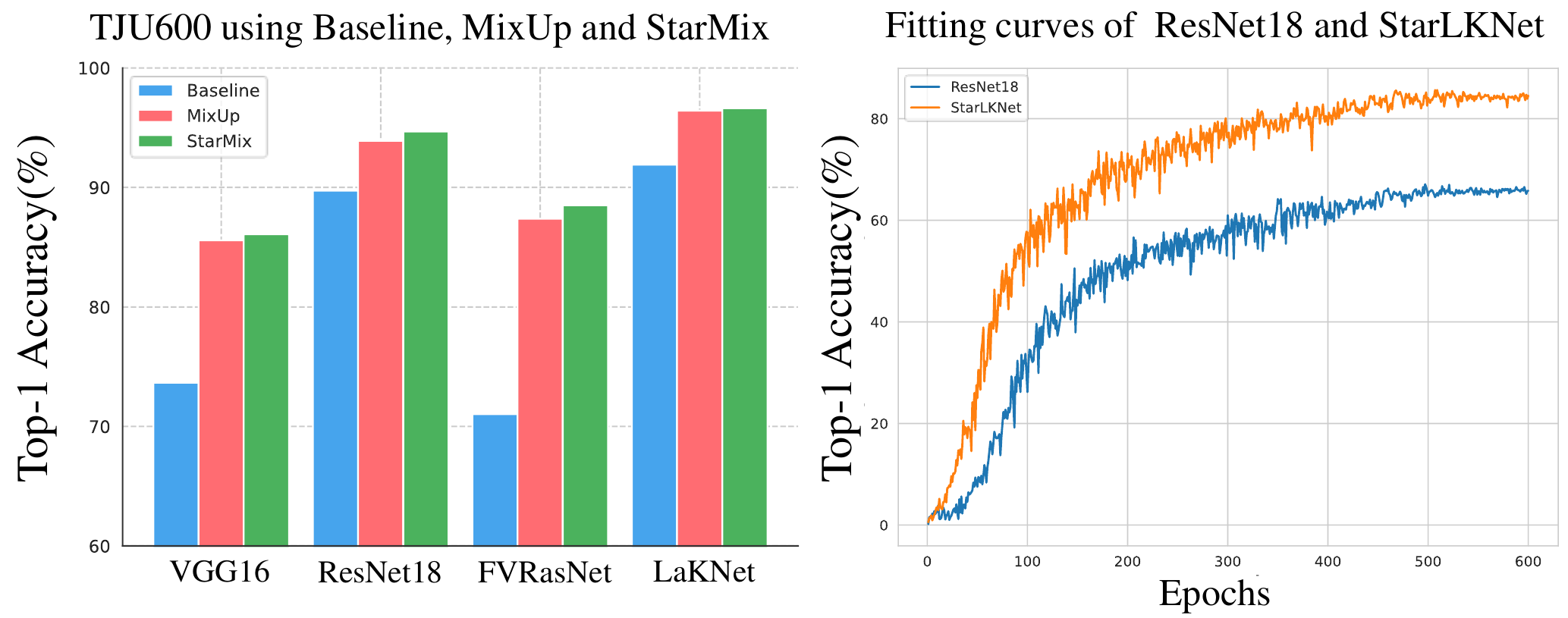}
    \caption{\emph{Left:} Top-1 Accuracy($\uparrow$) using MixUp and StarMix in different models; \emph{Right:} StarLKNet and ResNet18 fitting curves on the VERA220 dataset; StarLKNet fits faster and with higher classification accuracy.}
\end{wrapfigure} 
that the vein features' distribution is continuous and sparse. Some small convolutional kernel networks can't capture more features, which limits the performance of the model.
Some recent studies \cite{Ding2022replknet, Liu2022slak, Yu2024inceptionnext, Ding2024unireplknet} have indicated that with some subtle improvements, CNNs with a larger effective sense field can rival the performance of Vision Transformer (ViT) \cite{Dosovitskiy2021vit}. We found that large convolutional kernel networks have a significant advantage over small ones in capturing vein features' distribution. As shown in Figure 2, the large kernel exhibits faster fitting speeds and higher classification accuracies than the small kernel framework represented by ResNet18 \cite{He2016deep}.

To address the challenges of feature extraction from vein images, we propose a mixup-based method for vein images. Furthermore, we design a high-performance palm vein identification network based on the large convolutional kernels, to extract more comprehensive and robust feature representations of palm vein images. First, we propose StarMix, featuring a more suitable mask for vein image mixing than other mixup methods, with the mask generated from the Gaussian function. In our method, the network is free to choose the mixing strategy as either StarMix or vanilla mixup with a threshold. Furthermore, we propose LaKNet, a network with convolutional and gating modules. The convolutional module comprises large kernel convolution and small kernel convolution, employed for extracting global and local features. The gating module facilitates feature filtering by learning to control the flow of feature information. Experiment results show that the StarLKNet model exhibits superior performance compared to the ResNet18 model on the VERA220 \cite{Tome2015vera} dataset, with an improvement in the test set top-1 accuracy of \gbf{+19.73}\% without augmentation.

In summary, our main contributions are as follows:
\begin{itemize}
    \item We rethink the impact of convolutional kernel size on network performance in the vein identification task, and find that for vein images with continuous and sparse feature distributions, increasing the  \emph{Effective Receptive Field} can significantly improve network performance.
    \item We propose StarLKNet, a novel network framework designed for vein identification, that employs large convolutional kernels and a gating module to achieve comprehensive and robust feature extraction. Our evaluation of two large public palm vein datasets demonstrates that StarLKNet outperforms existing methods in terms of identification accuracy and validation error.
    \item We propose StarMix, a data augmentation method that utilizes a Gaussian function to generate, for mixing, suitable masks for vein image feature distribution, thereby significantly enhancing classifier performance.
\end{itemize}

\section{Related Work}\label{sec2}

\subsection{Palm Vein Identification} 
As Hemoglobin is absorbed in the infrared spectrum, infrared cameras can acquire vein images. Such an acquisition method and the characteristics of vein distribution bring challenges for feature extraction. Research made to address this challenge can be broadly categorized into two types of traditional methods, handcrafted features extracted and input to shallow machine learning, and CNN-based feature extraction methods: 1) In the first category, Miura \emph{et al.} \cite{Miura2007extractionfv}, for instance,  performed repeated line tracking to detect valley shapes in cross-sectional vein patterns and extract finger-vein texture for verification. Other works, assuming that vein patterns in a predefined neighbor can be regarded as line segments, have proposed line detection methods to extract line-like textures, including Gabor-based and wide line detectors \cite{Huang2010line}. Traditional shallow machine learning methods used in the vein domain, for instance, employed principal component analysis (PCA) and linear discriminant analysis (LDA) for feature dimensionality reduction, combined with support vector machine (SVM) for feature classification; 2) CNNs demonstrated remarkable capability in extracting features in vein identification tasks. Syafeeza \emph{et al.}, for instance, proposed a CNN with four layers for finger-vein identification \cite{Syafeeza2016FingerveinBI} and later employed a pre-trained VGG16 \cite{Simonyan2014vgg} model and an enhanced Conv with seven layers to identify the finger-veins  \cite{Liu2017FingerVR}.

\subsection{MixUp}
The inception of mixing augmentation methods started with MixUp \cite{Zhang2018mixup}, which consists of a static linear interpolation of two samples according to a mixing ratio from 0 to 1 to obtain a mixed sample. CutMix \cite{Yun2019cutmix} converts the MixUp sample from image pixels to space level, generating a mixing ratio-sized mask to randomly mix patches. Subsequently, several methods were proposed to improve the sample mixing policies or label mixing policies  \cite{Verma2019manifold, Qin2020resizemix, Li2021samix, Liu2022decoupledmix}. In contrast to hand-crafted methods, SaliencyMix \cite{Uddin2020saliencymix} obtains saliency information through an additional feature extractor, with guided mixing of the samples. With a similar aim, PuzzleMix \cite{Kim2020puzzle} and Co-Mix \cite{Kim2020co} utilize gradient information for backward propagation to locate feature regions and employ an optimal transport scheme to avoid overlapping feature information by feature maximization in the mixed samples. AutoMix \cite{Liu2022automix} adopts an end-to-end way by designing a generator mix block that optimizes both the generator and the model, achieving an optimal result in terms of time overhead and performance. More recently, AdAutoMix \cite{Qin2023adautomix}, built upon AutoMix, has been proposed to augment the generated samples by mixing any set of $N$ samples and not only two, and also proposed adversarial training to prevent generator overfitting, by pushing the generator to generate difficult samples with more impact on improving training performance.

\section{Preliminaries}\label{sec3}

\subsection{Mixup}
\label{sec:3.1}
We define $\mathbb{X}$ to be the set of training samples and $\mathbb{Y}$ the set of ground truth of the corresponding labels. For each sample pair $(x, y)$, $x \in \mathbb{R}^{w,h,c}$, and $y \in \mathbb{R}^C$ is the corresponding one-hot label. where $w,h,c$ are the sample's width, length, and channel, respectively; $C$ is the number of sample classes. We mix the sample pairs $(x_i,y_i)$, $(x_j,y_j)$ by linear interpolation according to MixUp to obtain mixed samples and labels:
\begin{equation}
    \begin{aligned}
    \label{eq:1}
        \hat{x} = \lambda * x_i + (1-\lambda) * x_j, \\
        \hat{y} = \lambda * y_i + (1-\lambda) * y_j,
    \end{aligned}
\end{equation}
where $\lambda$ is the mixing ratio from the $Beta(\alpha, \alpha)$ distribution. We map the mixed sample $\hat{x}$ to its label $\hat{y}$ by a deep neural network $f_\theta$: $f_\theta(\hat{x}) \mapsto \hat{y}$. $f$ trains the network vector parameter $\theta$ continuously by minimizing the loss function, \emph{i.e.}.

\subsection{Gating}
\label{sec:3.2}
The MogaNet \cite{Li2022MogaNet} gating mechanism in our network comprises $\rm{Conv}_{1\times1}$ and $\rm{SiLU}$ activation function. It provides a simple yet effective structure for filtering the flow of information during network training, thereby enhancing the efficacy of the extracted features. The objective of utilizing $\rm{SiLU}$ is to integrate the smoothness and nonlinearity of the $\rm{Sigmoid}$ function with the linear properties of the ReLU linear unit within the positive region:
\begin{equation}
    \begin{aligned}
    \label{eq:2}
        \rm{SiLU}(z) = z \cdot \rm{Sigmoid}(z).
    \end{aligned}
\end{equation}

Specifically, $\rm{Conv}_{1\times1}$ is employed to implement the linear transformation of features, whereby the weights are adjusted to emphasize or suppress the features. The output of $\rm{Conv}_{1\times1}$, $z \in \mathbb{R}^{w,h,c}$, is utilized as input to $\rm{SiLU}$, which is capable of adjusting its activation value according to the importance of the input features, thereby realizing the effect of the gating mechanism:
\begin{equation}
    \begin{aligned}
    \label{eq:3}
        \rm{gating}(z) = \rm{SiLU}(\rm{Conv}_{1\times1}(z)).
    \end{aligned}
\end{equation}

\section{StarLKNet}\label{sec4}

Our proposed StarLKNet model, illustrated in Figure \ref{fig: StarMix}, consists of two components, StarMix and LaKNet. StarMix employs a mask generated by a Gaussian function to mix and augment the data, while LaKNet consists of a convolutional module with a large kernel and a gating module. In the next section, we first introduce our mixup method StarMix, and then detail LaKNet.

\begin{figure}[t]
    \centering{
    \includegraphics[width=1.0\linewidth]{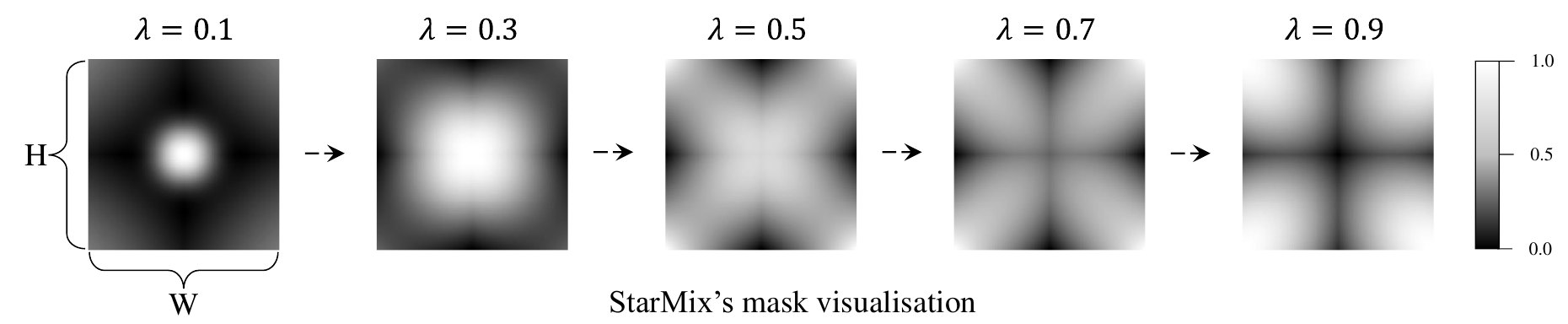}}
    \caption{From \emph{left} $\rightarrow$ \emph{right} are the visualizations of StarMask with different $\lambda$.}
    \label{fig: StarMask}
\end{figure}

\subsection{StarMix}
\label{sec:4.1}
This section details the StarMix method; the pseudo-code for StarMix is provided in Algorithm \ref{al:1}.

\paragraph{StarMask for Augmentation}  
As shown in Figure \ref{fig: StarMask}, compared to $\lambda$ in the vanilla MixUp, the mixing mask StarMask, focusing more on the surroundings and center of the samples, adapts well to the distribution of the vein features. To obtain StarMask, we define a Gaussian function $\rm{Ga}(\cdot)$ to generate the mask $G$. We assume a pair of samples $(x_i,y_i)$, $(x_j,y_j)$, with sample $x \in \mathbb{R}^{w,h,c}$, and we consider a mixing parameter drawn $\lambda$ from $\rm{Beta}(\alpha, \alpha)$. The Gaussian function is given by Eq.\ref{eq:4}:
\begin{equation}
\label{eq:4}
    \rm{Ga}(k, w, h, \sigma)=\mathrm{e}^{-\frac{\left(x_{G}-\frac{k}{2}\right)^{2}}{2 \sigma^{2}}},
\end{equation}
where $k$ is the Gaussian kernel`s size, set to 224, $x_G$ is the region division of the mask, $x_G=[k/2:k-k/2, \quad k/2:k/2-k]$, $\sigma = \lambda * h$. Passing the parameters to $\rm{Ga}(\cdot)$ as in Eq.\ref{eq:5} allows us to obtain the mask $\mathcal{M}$:
\begin{equation}
\label{eq:5}
    \mathcal{M} = \frac{1}{N} \sum_{n=1}^{N} Ga(k_n, h, \sigma_n),
\end{equation}
where $N = 3$. Note that $\sigma_2 = (1-\lambda)*h$, $\sigma_1 = \sigma_3$, and $k_3 = 2k_1 = 448$. We then normalize $\mathcal{M}$ to get the final mask $G$ as Eq.\ref{eq:6}:
\begin{equation}
\label{eq:6}
    G = \frac{\lambda}{1 + e^{-\mathcal{M}}}.
\end{equation}

\begin{algorithm}[t]
    \caption{StarMix pseudo-code process.}
    \begin{algorithmic}[1]
    \Require $\rm{Beta}(\alpha, \alpha)$ distribution, training samples and labels $\mathbb{X}$, $\mathbb{Y}$, mixing ratio $\lambda$, threshold [0.3, 0.7], Gaussian function $\rm{Ga}(\cdot)$ and $k$ is kernel size.
    \State $x \in \mathbb{R}^{w,h,c}$
    \While{$x_i$, $y_i$ in $\mathbb{X}$, $\mathbb{Y}$ loder}
    \State $\lambda$ = $\rm{Beta}(\alpha, \alpha)$,
    \State $x_j$, $y_j$ = torch.randperm($x_i$, $y_i$),
    \If{0.3 $\le \lambda \le$ 0.7} 
    \State $G$ = $\rm{Ga}(k, h, \sigma),$
    \State $\hat{\lambda}$ according to the Eq.\ref{eq:7},
    \State $\hat{x} = G * x_i + (1-G) * x_j$, 
    \State $\hat{y} = \hat{\lambda} * y_i + (1-\hat{\lambda}) * y_j$.
    \Else
    \State $\hat{x} = \lambda * x_i + (1-\lambda) * x_j$, 
    \State $\hat{y} = \lambda * y_i + (1-\lambda) * y_j$.
    \EndIf
    \EndWhile
    \end{algorithmic}
    \label{al:1}
\end{algorithm}
After we get the mask $G$, we need to recalculate the correct ratio for the mixing ratio $\lambda$ according to Eq.\ref{eq:7}:
\begin{equation}
\label{eq:7}
    \hat{\lambda} = \frac{ {\textstyle \sum_{i=1}^{w}}\textstyle \sum_{j=1}^{h}G_{i,j}}{w \times h},
\end{equation}
where $G_{i,j}$ denotes the pixel value at the \emph{i}th row and \emph{j}th column in $G$. $G_{i,j}$ can be seen as the average of all pixel values in $G$ in terms of intensity. Finally, the mixed sample $\hat{x}$ and the corresponding label $\hat{y}$ are obtained according to Eq.\ref{eq:8}:
\begin{equation}
\begin{aligned}
\label{eq:8}
    \hat{x} = G * x_i + (1-G) * x_j, \\
        \hat{y} = \hat{\lambda} * y_i + (1-\hat{\lambda}) * y_j.
\end{aligned}
\end{equation}

\paragraph{Threshold setting operation}
As shown in Figure \ref{fig: StarMask}, the obtained masks are not reliable when $\lambda$ is less than 0.3 or more than 0.7 due to $\rm{Ga}(\cdot)$, so we propose the use of thresholding to avoid obtaining unreliable masks when mixing:
\begin{equation}
\label{eq:9}
\begin{aligned}
\hat{x} & = \left\{\begin{matrix}
 G * x_i + (1-G) * x_j, & \rm{if} \quad 0.3 \le \lambda \le 0.7, \\
 \lambda * x_i + (1-\lambda ) * x_j, & \rm{otherwise}.
\end{matrix}\right.
\end{aligned}
\end{equation}
When $\lambda$ is within the [0.3, 0.7] threshold range, StarMix is used, and when $\lambda$ is outside it, we choose vanilla Mixup. This threshold setting avoids the disadvantages of StarMix in special cases. The benefits of mixing StarMix and vanilla Mixup can be seen in Table \ref{tab:starmix_ablation}.
\begin{figure*}[t]
    \centering
    \includegraphics[width=1.0\linewidth]{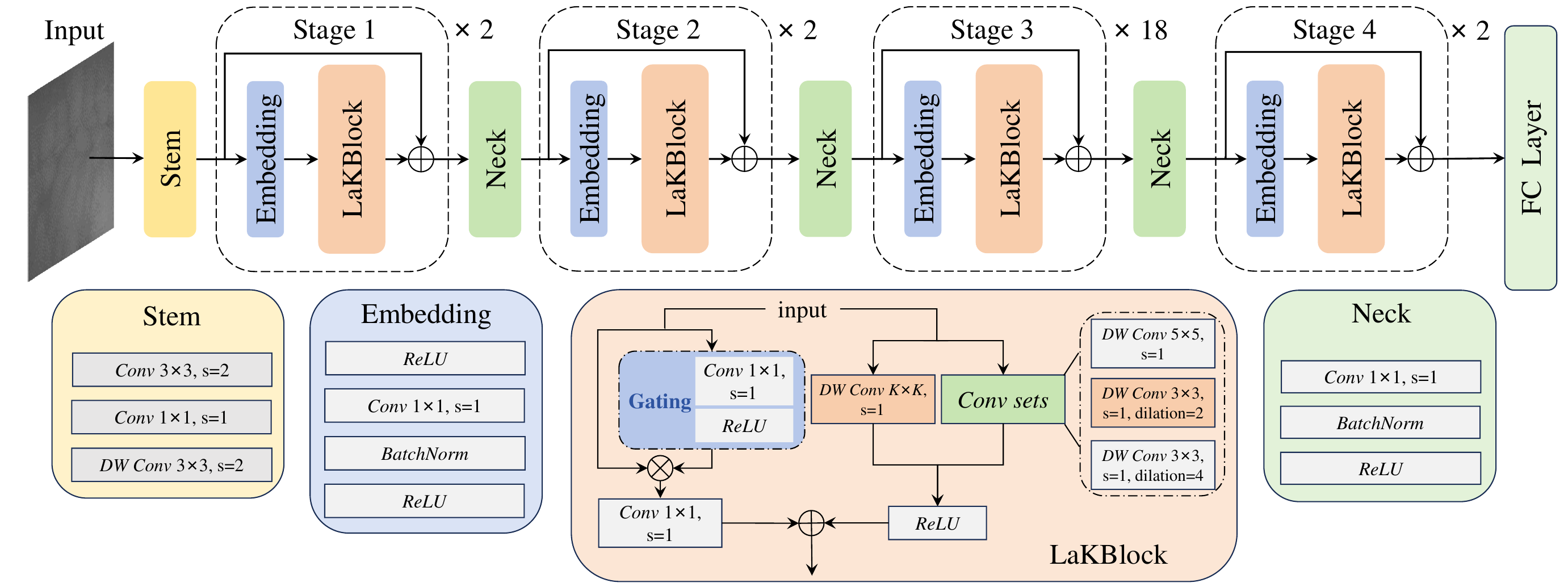}
    \caption{The framework of the proposed LaKNet. $Upper:$ Denotes the main module of LaKNet, containing 1 Stem, 4 Stages, 3 Necks, and an FC layer. $Lower:$ a comprehensive illustration of the functions and specific operations within each module.}
    \label{fig:LaKNet}
\end{figure*}
\subsection{LaKNet}
\label{sec:4.2}
This subsection we will presents a detailed description of the LaKNet modules and methods.

\paragraph{Architecture Specification}
\textbf{Stem} is the first module used as input to the network to capture more details by mapping the original samples into a higher dimensional space. Stem comprises a $\rm{Conv}_{3\times3}$ with step=2 and channel=64. It is then downsampled by a $\rm{Conv}_{1\times1}$ with step=1 and a $DW$ $\rm{Conv}_{3\times3}$ (Depthwise Separable Convolution) with step=2.

\textbf{Embedding} sub-module of the stage is comprised of a $\rm{Conv}_{1\times1}$, and a BatchNorm layer. The inputs and outputs of this module are activated by the ReLU function. The main purpose of Embedding is to aggregate further the feature information of Stage's inputs in the convolution, to prevent the model from overfitting the training set, and to avoid gradient exploding or vanishing. This is a common practice for most network models.

\textbf{LaKBlock} comprises two branches: a Conv module and a gating module. The Conv module employs a combination of $DW$ Conv with a large kernel and a Conv $set$, to extend the \emph{Effective Receptive Field} by capturing both global and local features. The Conv set uses some small convolution kernels, $\rm{Conv}_{5 \times 5}$, to capture local features, and two dilation convolutions aiming at capturing the discrete features without additional overheads. The gating module employs a gating mechanism formed by the combination of $\rm{Conv}_{1\times1}$ and ReLU activation functions to get effective features with input $\mathcal{Z}$. We finally merge the output of the Conv module and the output of the Gating module as the total output, which is used as input to the \textbf{Neck} layer.

\textbf{Neck} module position is used to connect different stages in the architecture and is used by a Conv$_{1\times1}$ to increase the dimension of the channel input for the next stage.

\textbf{FC Layer} module is a Linear layer intended to be used for the final classification, where the obtained feature information is subjected to a mapping to obtain the final classification probability distribution.

To summarize, the channel dimensions of each stage are distinct and the large convolutional dimensions are different due to the downsampling operations. Consequently, the number of layers, channel dimensions, and large convolutions of each stage in LaKNet are $\left\{2, 2, 18, 2\right\}$, $\left\{128, 256, 512, 1024\right\}$, and $\left\{31, 29, 27, 13\right\}$, respectively.

\paragraph{Kernels mixing.} This paragraph explains the large kernel function Lak$(\cdot)$ and a set of small kernel function Sak$(\cdot)$ sets mixing method, aiming to enable the model to capture both local and global features. This allows for stable optimization of the model $f_\theta$.
\begin{equation}
\label{eq:10}
    \mathcal{K} = \rm{Lak}(\mathcal{Z}) + \rm{Sak}(\mathcal{Z}),
\end{equation}
where $\mathcal{K}$ denotes the embedding layer output, $\mathcal{Z}$ was the Lak$(\cdot)$ and Sak$(\cdot)$ add output, Sak$(\cdot)$ was combined with a depthwise Conv$_{5\times5}$ and two depthwise $Dilation$ Conv$_{3\times3}$ with dilation set to 2 and 4.

\paragraph{Gating.} The Gating operation aims at emphasizing or suppressing features by learning the weights of Conv$_{1\times1}$, combined with the activation function to introduce nonlinearity and complete the feature screening process. The rationale behind selecting the faster ReLU over the SiLU is that the input $\mathcal{Z}$ is activated by ReLU before it is fed into LaKBlock, thus eliminating the need for additional negative semiaxis nonlinear features according to Eq.\ref{eq:11}:
\begin{equation}
    \begin{aligned}
    \label{eq:11}
        \rm{gating}(\mathcal{Z}) = \rm{ReLU}(\rm{Conv}_{1\times1}(\mathcal{Z})).
    \end{aligned}
\end{equation}

\section{Experiments}\label{sec5}

To evaluate the efficacy of our method, we conducted experimental comparisons on 2 palm vein datasets: TJU600 \cite{Zhang2018palmprint} and VERA220 \cite{Tome2015vera}, and 3 general datasets: CIFAR10 \& CIFAR100 \cite{Krizhevsky2009learning}, and CUB200 \cite{Wah2011caltech}. For vein identification, 6 benchmarks were used in 3 cases: baseline, MixUp, and StarMix. For fair validation, we compared some mainstream networks, \textit{i.e.} VGG16, ResNet18, ResNet50, and VanillaNet \cite{Chen2024vanillanet}; We also compared three classifiers in the vein task: FVCNN \cite{Das2018fvcnn}, PVCNN \cite{Qin2021pvcnn}, and FVRasNet \cite{Yang2020fvrasnet}. 
We refer to the median of the last 10 epochs of the test set for Top-1 accuracy as the final results, as the model's classification performance in that state tends to stabilize, which is a better measure of the model's final performance and the overfitting degree.
The False Acceptance Rate (FAR) and False Rejection Rate (FRR) of the last epoch are considered for computing the Equal Error Rate (EER), which is the error rate at a specific threshold when FAR and FRR become equal. Lower EER values indicate better verification performance.
We marked the best and second results using \textbf{bold} and \bb{cyan}.

\subsection{Dataset Information}
\label{sec:5.1}
We choose two large public palm vein datasets and three general classification datasets for our experiments:

\textbf{TJU600:} The TJU palm vein dataset consists of palm vein images provided for the left and right hands of 300 volunteers. The sample data from each volunteer was collected 2 times, with a time interval between the 2 collections of about 60 days. As 10 vein images were collected from each palm each time, the dataset contains 12,000 images (300 volunteers $\times$ 2 palms $\times$ 10 images $\times$ 2 time periods). The resolution of each image was scaled to 224 $\times$ 224 after ROI normalization.

\textbf{VERA220:} The VERA palm vein dataset contains palm vein images of the left and right hands of 110 volunteers, acquired over two periods. As 5 palm vein images were collected for each hand at each time, the dataset contains 2200 images (110 volunteers $\times$ 2 palms $\times$ 5 images $\times$ 2 time periods). The resolution of each image was also set to 224 $\times$ 224.

\textbf{CIFAR10 \& CIFAR100:} The CIFAR10 \& CIFAR100 datasets contain a total of 60,000 images, with 50,000 training images and 10,000 test images in 32 $\times$ 32 resolutions. CIFAR10 has 10 classes and CIFAR100 has 100 classes.

\textbf{CUB200:} CUB200 contains 11,788 images from 200 wild bird species. We followed the OpenMixup setting and divided them into train and test sets.

\subsection{Implementation Details}
\label{sec:5.2}
For TJU600, we divided the total dataset into a training set and a test set, each consisting of 6000 images (300 volunteers $\times$ 2 palms $\times$ 10 images). Similarly, we divided the VERA220 dataset into a training set and a test set, each consisting of 1100 images (110 volunteers $\times$ 2 palms $\times$ 5 images). We use RandomFlip and padding 3 pixels of RandomCrop as the basic augmentation methods. The batch size of the experiments was set to 32, and a total of 600 epochs were considered for training; the learning rate was set to 0.1. For VGG16, the learning rate was set to 0.01, adjusted by the cosine scheduler, For the FVCNN and PVCNN the Adam \cite{Ilya2019AdamW} optimizer with a learning rate was set to 0.0001, weight decay of 0.001; other models' training was performed by the momentum of 0.9 and a weight decay of 0.0001 by the SGD \cite{Loshchilov2016sgdr} optimizer. For MixUp and StarMix, the hyperparameter Beta$(\alpha, \alpha)$, $\alpha$ = 1. 

For the experiment settings of StarMix in the general dataset classification task, we followed the settings in OpenMixup \cite{Li2022openmixup}. For the CIFAR10 and CIFAR100 datasets, we used the ResNet18 and ResNeXt50 trained with 800 epochs, learning rate of 0.1, and the SGD optimizer with momentum of 0.9, weight decay of 0.0001, dynamically adjusted by cosine scheduler, and batch size of 100. For the CUB200 dataset, we use the official PyTorch pre-trained ResNet18 and ResNet50 models on ImageNet-1k are adopted as initialization, trained with 200 epochs, using SGD optimizer with momentum of 0.9, weight decay of 0.0005, batch size of 16, learning rate of 0.001, dynamically adjusted by cosine scheduler.
\begin{table}
\centering
\setlength{\tabcolsep}{4mm}
\caption{Top-1 accuracy(\%)$\uparrow$ and EER(\%)$\downarrow$ of Baseline, MixUp,and StarMix on TJU600.}
    \begin{tabular}{c | c c | c c | c c}
    \toprule
    \multirow{2}*{\textbf{TJU600}} &\multicolumn{2}{c|}{Baseline} &\multicolumn{2}{c|}{with MixUp} &\multicolumn{2}{c}{with StarMix} \\
      & Acc $\uparrow$ & EER $\downarrow$ & Acc $\uparrow$ & EER $\downarrow$ & Acc $\uparrow$ & EER $\downarrow$ \\
    \hline
    VGG16       & 73.65  & 4.07 & 86.08 & 2.6 & 85.57 & 2.52 \\
    
    ResNet18    & \bb{89.73}  & \bb{1.2} & 93.90 & 0.81 & 94.67 & \bb{0.71} \\
    
    ResNet50    & 85.68  & 1.96 & 92.87 & 0.95 & 94.13 & 0.93 \\

    FVCNN       & 58.65  & 9.02 & 79.19 & 3.73 & 80.45 & 3.58 \\
        
    FVRASNet    & 71.02  & 4.19 & 87.38 & 2.11 & 88.50 & 2.01 \\
    
    PVCNN       & 62.57  & 6.06 & 86.38 & 2.46 & 86.92 & 2.36 \\

    VanillaNet  & 86.56  & 2.0 & \bb{94.87} & \bb{0.78} & \bb{95.75} & 0.72 \\
    
    Ours  & \textbf{91.90}  & \textbf{1.06} & \textbf{96.43} & \textbf{0.51} & \textbf{96.63} & \textbf{0.44} \\ \hline
    
    Gain & \gbf{+2.17}  & \gbf{-0.14} & \gbf{+1.55} & \gbf{-0.27} & \gbf{+0.88} & \gbf{-0.27} \\
    \bottomrule
    \end{tabular}
\label{tab:tju600}
\end{table}

\begin{table}
\centering
\setlength{\tabcolsep}{4mm}
\caption{Top-1 accuracy(\%)$\uparrow$ and EER(\%)$\downarrow$ of Baseline, MixUp,and StarMix on VERA220.}
    \begin{tabular}{c | c c | c c | c c}
    \toprule
    \multirow{2}*{\textbf{VERA220}} &\multicolumn{2}{c|}{Baseline} &\multicolumn{2}{c|}{with MixUp} &\multicolumn{2}{c}{with StarMix} \\
       & Acc $\uparrow$ & EER $\downarrow$ & Acc $\uparrow$ & EER $\downarrow$ & Acc $\uparrow$ & EER $\downarrow$ \\ \hline
    
    VGG16      & 50.82  & 14.35 & 90.45 & 1.53 & 89.27 & 1.26 \\
    
    ResNet18   & \bb{65.82}  & 7.86 & 94.82 & 1.12 & 95.73 & \bb{0.61} \\
    
    ResNet50   & 56.18  & 9.74 & 87.82 & 2.25 & 91.73 & 1.56 \\

    FVCNN      & 61.91  & 6.64 & 78.64 & 3.41 & 79.09 & 3.50 \\
    
    FVRASNet   & 61.73  & 7.74 & 83.36 & 2.36 & 89.09 & 2.28 \\
    
    PVCNN      & 50.82  & 11.42 & 89.55 & 1.8 & 89.73 & 1.19 \\

    VanillaNet & 61.80  & \bb{7.42} & \bb{97.09} & \bb{0.98} & \bb{97.27} & 0.64 \\
    
    Ours &\textbf{85.55}  &\textbf{1.81} &\textbf{97.52} &\textbf{0.51} &\textbf{98.09} &\textbf{0.35}  \\ \hline
    Gain & \gbf{+19.73}  & \gbf{-5.56} & \gbf{+0.43} & \gbf{-0.47} & \gbf{+0.82} & \gbf{-0.26} \\
    \bottomrule
    \end{tabular}
\label{tab:vera220}
\end{table}

\subsection{Classification Results}
\label{sec:5.3}
Table \ref{tab:tju600} and Table \ref{tab:vera220} show the performance of StarMix and LaKNet in terms of 2 metrics: classification accuracy and EER. Table \ref{tab:tju600} shows that our method outperforms the existing methods on TJU600 for classification performance. StarMix achieves good gains in different models; in particular, in ResNet50 and FVCNN, we achieve \textbf{94.13}\% and \textbf{80.45}\% and outperform vanilla Mixup by \gbf{+1.26}\% and \gbf{+1.26}\%. Compared to other models, StarLKNet achieves \textbf{91.90}\% accuracy without the augmentation method and again \textbf{96.63}\% with StarMix. Similarly, the results of our method on VERA220, are shown in Table \ref{tab:vera220}. We observe that due to the small VERA220 dataset size, the performance of the model without augmentation is minimal. After using Mixup and StarMix, the performance improved significantly. VGG16 and PVCNN, for instance, get an improvement of \gbf{+39.63}\% and \gbf{+39.18}\%, respectively. But in the same case, without using augmentation methods, StarLKNet gets also a good result, which favorably demonstrates the ability of StarLKNet for effective feature extraction.

\subsection{Robustness}
\label{sec:5.4}

\paragraph{Equal Error Rate}
\label{sec:5.4.1}
\begin{figure}[b]
    \centering
    \includegraphics[width=1.0\linewidth]{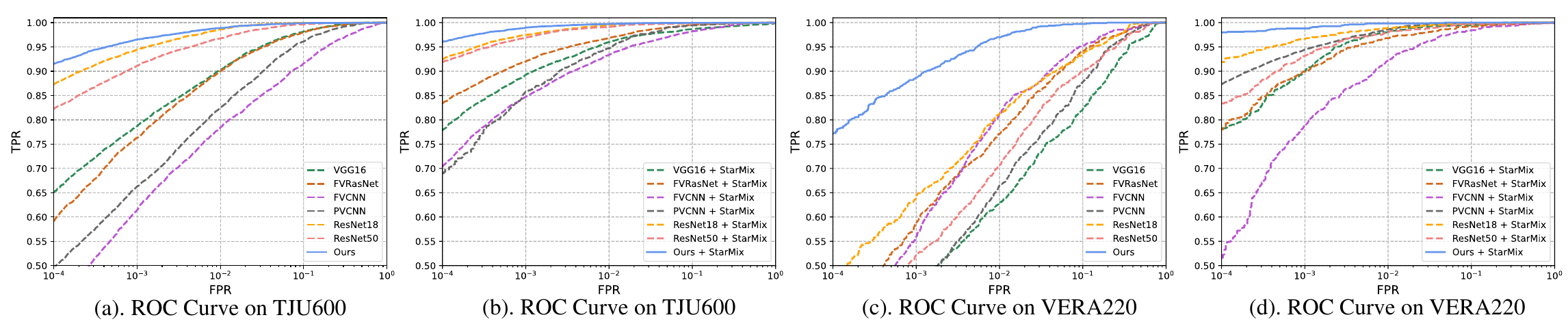}
    \vspace{-10pt}
    \caption{\textbf{(a).} shows the ROC curves for different models on TJU600; \textbf{(b).} shows the ROC curves for different models using StarMix on TJU600; \textbf{(c).} shows the ROC curves for different models on VERA220; \textbf{(d).} shows the ROC curves for different models using StarMix on VERA220.}
    \label{fig:5}
\end{figure}
In the biometric identification task, positive and negative samples are usually highly imbalanced, and there is a clear trade-off relationship between FAR and FRR. Consequently, identification technology necessitates rigorous criteria for model performance evaluation. The conventional single index of accuracy is insufficient to reflect model performance and stability fully. The ROC (Receiver Operating Characteristic) curve and the EER offer a comprehensive evaluation of a biometric model's performance in practical applications.

In the four subfigures of Figure.\ref{fig:5}, we observe that our method achieves the best ROC curves both with and without StarMix, and it can also be seen that the gain for the model with the use of StarMix is significant.

\paragraph{Occlusion Robustness}
\label{sec:5.4.2}
\begin{wrapfigure}{r}[0cm]{0pt}
    \includegraphics[width=0.6\linewidth]{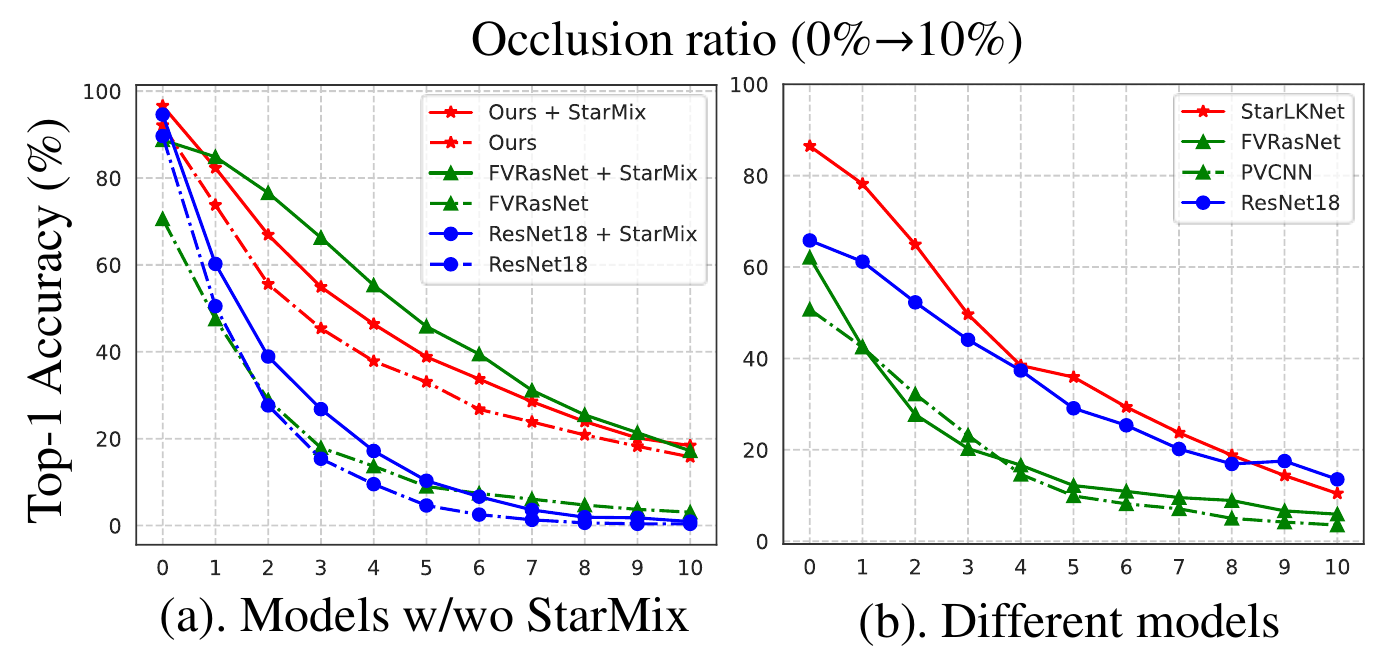}
    \caption{(a). shows the occlusion performance of models in TJU600 with and without StarMix. (b). shows the different models' performance about occlusion in VERA220.}
    \label{fig:6}
\end{wrapfigure}
To analyze the robustness of StarLKNet to random occlusions  \cite{Naseer2021occlusion}, we input the occluded test sets into different models for classification, test images were occluded by random 16 $\times$ 16 size patches, with the occlusion ratio gradually increasing from 0\% to 10\%. 
Since the features of vein images are continuous and discrete, a mask patch may cause the features to be interrupted when it falls in a region with constant veins. This is challenging to the identification model and offers an additional configuration to assess the model's robustness to vein features. 
From Figure \ref{fig:6}, we observe that StarLKNet achieves the highest accuracy for occlusion ratios from 0\% $\to$ 8\%, which shows that the large kernels are beneficial for robustness. Compared to the baseline, StarMix has also been shown to be beneficial for robustness.
\subsection{Analysis Study}
\label{sec:5.5}
In this subsection, we will discuss the effectiveness of the StarMix method and LaKNet architecture. \textbf{(\romannumeral1)} Does the StarMix can generalization for other datasets? \textbf{(\romannumeral2)} Does the larger kernel network capture more features in palm vein samples? 

\begin{table}[b]
\centering
\setlength{\tabcolsep}{1.75mm}
\caption{Top-1 accuracy(\%)$\uparrow$ of mainstream mixup methods on general datasets. All the results are from OpenMixup and AdAutoMix.}
    \begin{tabular}{c | c | c c | c c | c c}
    \toprule
     \multirow{2}*{\textbf{Method}} & Beta &\multicolumn{2}{c|}{CIFAR10} &\multicolumn{2}{c|}{CIFAR100} &\multicolumn{2}{c}{CUB200} \\
     & $\alpha$  & ResNet18    & ResNeXt50    & ResNet18   & ResNeXt50  & ResNet18    & ResNet50 \\ \hline
    
    MixUp    & 1.0   & 96.62  & 96.23   & 79.12 & 82.10 & 78.39  & 82.98 \\
    
    CutMix   & 0.2   & 96.68  & 96.60   & 78.17 & 78.32 & 78.40  & 83.17 \\
    
    FMix     & 0.2   & 96.58  & 96.76   & 79.69 & 79.02 & 77.28  & 83.34 \\
    
    SmoothMix & 0.2  & 95.84     & 96.26      & 78.69 & 78.85 & 76.42  & 82.78 \\
    
    GridMix   & 0.2  & 96.42     & 96.74      & 78.72 & 78.90 & 77.15  & 81.98 \\
    
    SaliencyMix & 1.0 & 96.05     & 96.70      & 79.12 & 78.77 & 77.95  & 82.02 \\ \hline

    \emph{Avg.} & - & \underline{96.34}     & \underline{96.55}       & \underline{78.92}  & \underline{79.49}    & \underline{77.59}      & \underline{82.71}    \\
    
    StarMix & 1.0 &\textbf{96.68}  &\textbf{97.11} &\textbf{79.73} &\textbf{80.44} &\textbf{78.82} &\textbf{83.38}  \\
    \bottomrule
    \end{tabular}
\label{tab:general_results}
\end{table}

\paragraph{StarMix transfer on general datasets} 
As a mixup-based data augmentation method, we further explore the performance of StarMix on some general datasets in the classification task. We compare some mainstream hand-crafted mixup methods, \emph{e.g.}, SaliencyMix, CutMix, FMix \cite{Harris2020fmix}, SmoothMix \cite{Lee2020smoothmix} and GridMix \cite{Baek2021gridmix}. Table \ref{tab:general_results} shows our StarMix can transfer some general tasks. All the results we refer from OpenMixup \cite{Li2022openmixup} and AdAutoMix \cite{Qin2023adautomix}. Table \ref{tab:general_results} shows the results of our StarMix on the general dataset. It can be shown that StarMix can achieve the average results of some hand-crafted methods with some improvement.

\begin{figure}[htbp]
	\begin{minipage}{0.5\linewidth}
		\centering
            \includegraphics[width=1.0\linewidth]{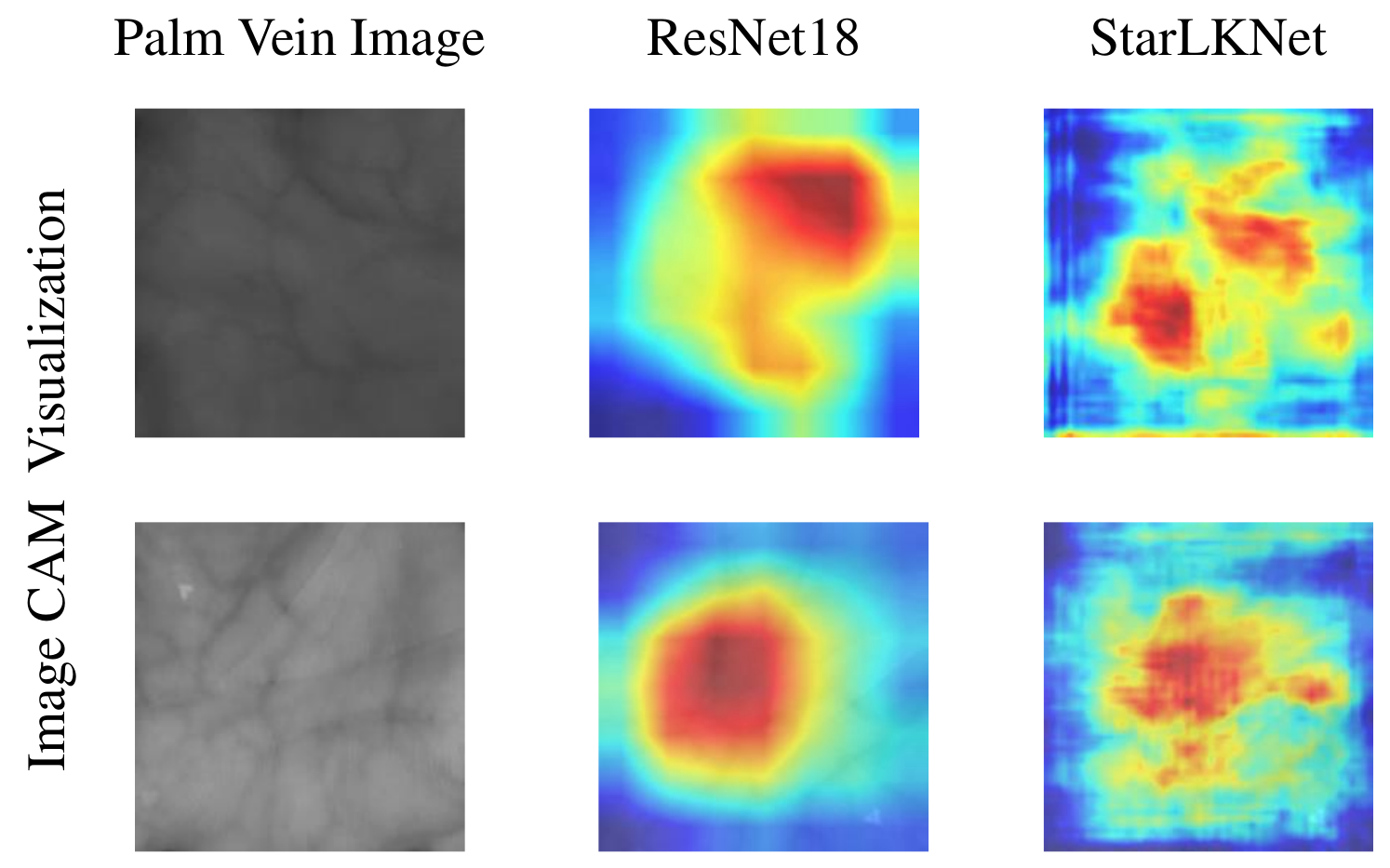}
            \vspace{-7pt}
            \caption{CAM visualization images of the small convolutional model ResNet18 and the large convolutional model LaKNet on the VERA220 dataset.}
            \label{fig: CAM}
	\end{minipage}
	\begin{minipage}{0.5\linewidth}
		\centering
            \includegraphics[width=1.0\linewidth]{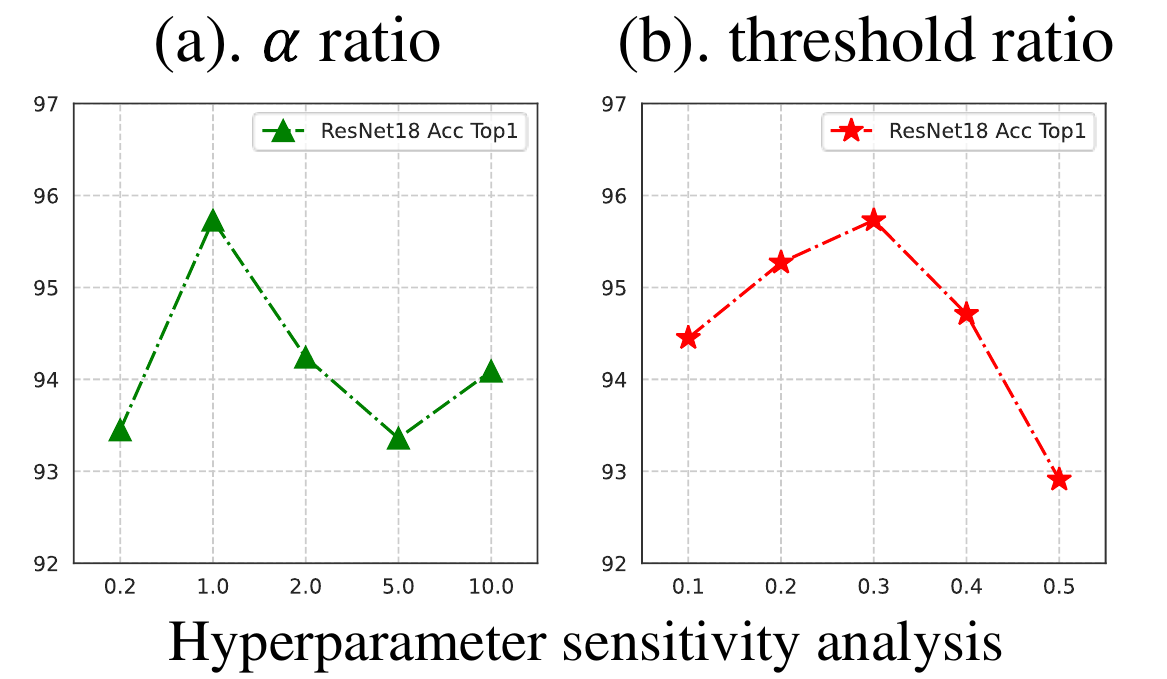}
            \caption{Two different hyperparameters identification Top-1 ($\uparrow$) results with different ratios on the VERA220 dataset.}
            \label{fig: hyper}
	\end{minipage}
\end{figure}

\paragraph{Visualization results} 
Figure \ref{fig: CAM} shows the Class Activation Mapping \cite{Selvaraju2017grad} (CAM) visualization images that our LaKNet can capture more feature information compared to the mainstream small convolutional network architecture ResNet18 and captures the distribution of vein texture in a more detailed way.

The top two images of Figure \ref{fig: CAM} show that ResNet18 captures only the approximate box of the vein distribution in one region of the image. In contrast, LaKNet additionally captures the vein distribution in another region. This not only demonstrates the performance of the large convolution module but also shows why our method gains a lot on the VERA220 dataset.

\subsection{Ablation Study}
\label{sec:5.6}
Our ablation experiments aim to analyze the effectiveness of the StarMix augmentation method, the Gating module, and the Large Kernel module in LaKNet. 
And for the hyperparameters of StarMix, \emph{e.g.}, Beta($\alpha, \alpha$) and threshold ratio analysis experiments are shown in Figure \ref{fig: hyper}.
\begin{itemize}
    \item We perform experiments using ResNet18 and FVRasNet on the VERA220 dataset. The aim is to verify the effect of StarMix and the [0.3, 0.7] threshold setting. Table \ref{tab:starmix_ablation} demonstrates that applying StarMix is detrimental to the model in certain instances, resulting in a 0.1\% reduction in Top-1 accuracy compared to the MixUp test set. As illustrated in Figure \ref{fig: StarMask}, the mask is observed to concentrate on the central region when $\lambda$ is less than 0.3 and on the periphery when it is greater than 0.7. However, when $\lambda$ is between 0.3 and 0.7, the mask achieves an additional boost by outperforming MixUp by \gbf{+0.91}\% and \gbf{+5.73}\% on ResNet18 and FVRasNet, respectively.
    \item In Table \ref{tab:laknet_ablation}, we have performed ablation experiments on each module of LaKNet to verify the effect of the corresponding module. It can be seen that the Conv set provides \gbf{+0.36}\% improvement compared to LaKNet with only Conv$_{K\times K}$ and a Conv$_{5\times5}$. The gating operation also helps the model's classification performance by \gbf{+0.74}\%. This is reasonable because the Conv set expands the \emph{Effective Receptive Field} using dilated Convolutions, which allows the model to capture more global features. The gating module filters the features, further retaining the valid high-dimensional features.
\end{itemize}
\begin{figure*}
    \begin{minipage}[t]{0.6\textwidth}
    \makeatletter\def\@captype{table}
        \centering
        \setlength{\tabcolsep}{1.0mm}
        \caption{Experiments about StarMix and threshold setting method on the VERA220 based ResNet18 and FVRasNet.}
            \begin{tabular}{c| c c c }
            \toprule
            VERA220      & ResNet18       & FVRasNet       & VanillaNet\\ \hline
            Baseline     & 65.82          & 61.73          & 61.80\\
            \emph{Mixup} & \textcolor{mygray}{94.82}   & \textcolor{mygray}{83.36}   & \textcolor{mygray}{97.09} \\
            StarMix      & 94.72          & 87.37          &  96.82\\
             w threshold & 95.73          & 89.09          &  97.27 \\ \hline
            Gain         & \gbf{+1.01} & \gbf{+1.72} & \gbf{+0.45} \\
            \bottomrule
            \end{tabular}
        \label{tab:starmix_ablation}
    \end{minipage}
    \begin{minipage}[t]{0.4\textwidth}
    \makeatletter\def\@captype{table}
        \centering
        \setlength{\tabcolsep}{1.0mm}
        \caption{Ablation experiments about LakNet and modules.}
            \begin{tabular}{c| c c }
            \toprule
            Modules            & TJU600 & VERA220 \\ \hline
            \emph{ResNet18}    & \textcolor{mygray}{89.73} & \textcolor{mygray}{65.82} \\
            LaKNet             & 89.78 & 84.45 \\
            + Conv set     & 90.12 & 84.81 \\
            + gating           & 91.90 & 85.55 \\ \hline
            Gain               & \gbf{+1.78} & \gbf{+0.74}  \\ 
            \bottomrule
            \end{tabular}
        \label{tab:laknet_ablation}
    \end{minipage}
\end{figure*}

\section{Conclusion}\label{sec6}
In this paper, we have proposed StarLKNet, a Conv-based palm vein identification network with large kernels, that incorporates the StarMix data augmentation method and the LaKNet architecture. StarMix generates masks for image mixing through the Gaussian function, which effectively reduces the overfitting problem caused by insufficient samples in the dataset and improves the model's robustness. The LaKNet network captures more globally effective features through large a \emph{Effective Receptive Field} combined with the screening capability of the gating mechanism, thus stabilizing the identification capability of the model. A series of classification and analysis experiments have been conducted to validate the outstanding performance of our method.

\bmhead{Acknowledgements}

This work was supported in part by Project of CQ CSTC (Grant No. cstc2018jcyjAX0057), in part by Science and Technology Research Program of Chongqing Municipal Education Commission (Grant No. KJQN201800814), in part by Opening Foundation of Chongqing Technology and Business University (Grant No. KFJJ2019103), and in part by Innovative Research Projects for Postgraduate Students of Chongqing Technology and Business University (No. yjscxx2024-284-53, No. yiscxx2024-284-231, No. yjscxx2024-284-234).


\bibliography{StarLKNet}



\end{document}